\documentclass[dvipsnames]{article}

\usepackage[nonatbib, final]{neurips_2022}
\usepackage[utf8]{inputenc} 
\usepackage[style=numeric, maxbibnames=99]{biblatex}
\usepackage[T1]{fontenc}    
\usepackage{hyperref}       
\usepackage{url}            
\usepackage{booktabs}       
\usepackage{amsfonts}       
\usepackage{amssymb}
\usepackage{nicefrac}       
\usepackage{microtype}      
\usepackage{xcolor}        
\usepackage{xspace}
\usepackage{bm}
\usepackage{amsmath}
\usepackage{graphicx}
\usepackage{multicol}
\usepackage{makecell} 
\newcommand{\name}{HARRIS\xspace}
\renewcommand{\vec}[1]{\boldsymbol{#1}}

\usepackage{ifthen}
\newboolean{blind}
\setboolean{blind}{false}

\title{HARRIS: Hybrid Ranking and Regression Forests\\for Algorithm Selection}

\ifthenelse{\boolean{blind}}
{
\author{
    Anonymous authors\\
}
}
{
\author{%
  Lukas Fehring$^1$, Jonas Hanselle$^1$, Alexander Tornede$^2$\\
  $^1$ Department of Computer Science, Paderborn University, Germany \\
  $^2$ Institute of Artificial Intelligence, Leibniz University Hannover \\
  \texttt{fehring2@mail.upb.de, jonas.hanselle@upb.de, tornede@tnt.uni-hannover.de}
}
}

\newcommand{\Algorithms}{\mathcal{A}}
\newcommand{\InstanceSpace}{\mathcal{I}}

\usepackage{todonotes}
\usepackage{multirow}
\newcommand{\dataset}{\mathcal{D}}
\newcommand{\nodewiseLoss}{\mathcal{L}}

\newcommand{\argmin}{\operatornamewithlimits{arg\,min}}

\begin{document}

\maketitle

\begin{abstract}
It is well known that different algorithms perform differently well on an instance of an algorithmic problem, motivating algorithm selection (AS): Given an instance of an algorithmic problem, which is the most suitable algorithm to solve it? As such, the AS problem has received considerable attention resulting in various approaches -- many of which either solve a regression or ranking problem under the hood. Although both of these formulations yield very natural ways to tackle AS, they have considerable weaknesses. On the one hand, correctly predicting the performance of an algorithm on an instance is a sufficient, but not a necessary condition to produce a correct ranking over algorithms and in particular ranking the best algorithm first. On the other hand, classical ranking approaches often do not account for concrete performance values available in the training data, but only leverage rankings composed from such data. We propose \name - Hybrid rAnking and RegRessIon foreSts - a new algorithm selector leveraging special forests, combining the strengths of both approaches while alleviating their weaknesses. \name' decisions are based on a forest model, whose trees are created based on splits optimized on a hybrid ranking and regression loss function. As our preliminary experimental study on ASLib shows, \name improves over standard algorithm selection approaches on some scenarios showing that combining ranking and regression in trees is indeed promising for AS.
\end{abstract}

\section{Introduction}
To this day, there are competitions on solving hard instances of the SAT (boolean satisfiability problem) problem \autocite{SAT2018,SAT2020}. In these competitions, one deals with a set of problems with the goal of solving them faster than the competitors. Here, the participants rarely use one algorithm to solve all problem instances. Instead, they utilize so-called algorithm selectors, often featuring machine learning models at their core, to predict the performance of different algorithms on the instance to select the one presumably performing best. In practice, most algorithm selectors either leverage a regression \cite{xu2007satzilla,AsLib,hanselleTWH21} or a ranking \cite{brazdil2000comparison,cunha2018label,tornedeWH20} model to predict the best algorithm. 

Unfortunately, both ranking and regression models feature considerable drawbacks when used at the core of a selector. While creating a ranking across algorithms according to their predicted performance does indeed yield the correct ranking as long as the predictions are correct, such a ranking can also be created without correctly estimating the performance. More precisely, correct performance predictions are a sufficient, but not a necessary criterion to create a correct ranking across the algorithms. Correspondingly, one may wonder whether solving a regression problem might not be much harder than what is required. From this perspective, ranking models are a more intuitive solution. However, they often do not take the concrete performance values, which are usually present as training data, into account, but are trained based on rankings created from these. Correspondingly, these ranking models are trained based on qualitative comparisons losing the actual quantitative information contained in the precise performance evaluations. As such, they lack the means to quantify how close two algorithms are in a predicted ranking and thus are more susceptible to problems arising from algorithms with actually very similar performance.

In this paper, we propose a new algorithm selector leveraging a machine learning model trained based on a composite loss with both a ranking and regression component, dubbed \name. In particular, the core of \name is formed by a random forest, whose trees are formed according to splits optimized on the aforementioned composite loss. By doing so, \name combines the strengths of both ranking and regression models while alleviating their weaknesses. 

\section{The Algorithm Selection Problem}
In Algorithm Selection (AS) \cite{rice_algorithm_selection}, we aim to find the best algorithm $A_i$ from a set of candidate algorithms $\{A_1,...,A_k\} = \Algorithms$ for a problem instance $I \in \InstanceSpace$ from a problem instance space $\InstanceSpace$. Formally, we seek to find a mapping, called algorithm selector $s:\InstanceSpace \rightarrow \Algorithms$, which maximizes a costly-to-evaluate performance measure $ m: \Algorithms \times \InstanceSpace \rightarrow \mathbb{R}$. Correspondingly, the optimal selector, called an oracle, is defined as 
\begin{equation}\label{math:oracle}
    s^*(I) \in \arg\max\limits_{A \in \Algorithms} \mathbb{E}[m(A, I)] \,\,\, .   
\end{equation}

As the performance measure $m$ is costly to evaluate, an exhaustive enumeration over the set of algorithms to choose the best performing one is no practical solution. This holds especially for constraint satisfaction problems, where one is finally interested in the solution to the instance, which is obtained as a result of the first algorithm run anyway. As a solution to this, most AS approaches leverage machine learning to learn a surrogate performance measure $\widehat{m}: \Algorithms \times \InstanceSpace \rightarrow \mathbb{R}$ mimicking the original performance measure $m$, while, in contrast to the original performance measure, being cheap to evaluate. Using such a surrogate $\widehat{m}$, selectors can be constructed as $ s(I) = \arg\max\limits_{A \in \Algorithms} \widehat{m}(A, I)$.

To learn such surrogates, we assume that we can represent instances in terms of features, which are at least somewhat correlated with the performance of one or multiple of the algorithms. Formally, these features are computed by a feature function $g:\InstanceSpace \rightarrow \mathcal{X}$ and we will write $\vec{x}_I \in \mathcal{X}$, when we want to address the features of instance $i \in \InstanceSpace$. When considering the algorithmic problem of SAT, such features could be, for example, the number of clauses or the number of variables. Moreover, we assume that we are given some prior evaluations of the performance measure $m$ for at least some of the algorithms on some training instances $\InstanceSpace_{\mathit{train}} \subset \InstanceSpace$, which we can use for learning. More formally, we assume training data with labels $\vec{y}_I = [m(I, A_1), \ldots, m(I, A_k)] \in \mathbb{R}^k$ where $A_i \in \Algorithms$, i.e.,
\begin{equation}\label{eq:dataset}
    \mathcal{D}_{\mathit{train}} = \left\{ (\vec{x}_I, \vec{y}_I) \vert I \in \InstanceSpace_{\mathit{train}}\right\} \,\,\, .
\end{equation}

\section{From Pure Ranking or Regression to Hybrid Ranking and Regression}
In practice, the surrogate performance measure $\widehat{m}$ is often implemented as a regression or ranking model based on a loss function $\ell: \mathbb{R}^k \times \mathbb{R}^k \rightarrow \mathbb{R}$, where we assume rankings to be represented as a $k$-dimensional real-valued vector for simplicity. While the first kind of models is trained using a regression loss function such as the mean squared error, which is aimed at minimizing the differences between the predicted algorithm performances $\widehat{m}(\cdot, \cdot)$ and the true performances $m(\cdot, \cdot)$ on the training data $\mathcal{D}$ making it a quantitative approach. Contrary to that, ranking models are trained based on ranking losses such as the (inverse of the) Spearman correlation \cite{spearman_general_1961}, which tries to maximize the correlation between the ranking across the algorithms imposed by the predicted latent utility values $\widehat{m}(\cdot, \cdot)$ and the ranking imposed by the true performances $m(\cdot, \cdot)$ making it a qualitative approach. 

Recall that both of these approaches have a significant disadvantage: On the one hand, regression approaches try to predict the performance of an algorithm on an instance as accurately as possible, solving a, perhaps, harder problem than necessary as we are actually just interested in correctly \textit{ranking} the algorithms. On the other hand, ranking approaches often ignore the concrete performance evaluations available in the training data and instead focus only on the ground truth ranking imposed by such values and correspondingly, ignore valuable data. 

This problem has been discussed before in \autocite{jonas_combined_ranking_and_regerssion} in the context of AS (and earlier in a more general setting in \cite{CombinedRankingAndRegression}), who advocate leveraging hybrid ranking and regression loss functions
\begin{equation}
    \ell_{\lambda}(\vec{y}, \vec{\widehat{y}}) = \lambda \ell_{\textit{regression}}(\vec{y}, \vec{\widehat{y}}) + (1- \lambda)\ell_{\textit{ranking}}(\vec{y}, \vec{\widehat{y}})
\end{equation} composed of a convex combination of a regression loss function $\ell_{\textit{regression}}: \mathbb{R}^k \times \mathbb{R}^k \rightarrow \mathbb{R}$ and a ranking loss function $\ell_{\textit{ranking}}: \mathbb{R}^k \times \mathbb{R}^k \rightarrow \mathbb{R}$. Here, $\lambda \in [0,1]$ is a hyperparameter controlling how strong the two loss functions influence the hybrid loss. The underlying idea is to leverage the strengths of the two approach classes, i.e., focusing on the ranking problem while also incorporating the precise performance information available in the training data and as such, eliminate their main weaknesses in the context of AS. The authors of \cite{jonas_combined_ranking_and_regerssion} found that training simple linear models and neural networks to predict latent utility values for algorithms based on such a hybrid loss function can indeed be beneficial and in particular, that values of $0 < \lambda < 1$ can yield the best performance.

\section{Hybrid Ranking and Regression Forests}
Building upon the successful work \cite{jonas_combined_ranking_and_regerssion}, in this work, we generalize the idea of training models based on such a hybrid loss function to tree-based models, known to be very effective in AS \cite{RunToSurvive}. We build forests of hybrid trees, detailed in the following, analogously to standard random forests \cite{breiman01}.

Recall that decision trees \autocite{breimann_cart} are trained by splitting the training data $\dataset_{train}$ recursively into two subsets, i.e., nodes $\dataset^+_{train}, \dataset^-_{train}$ based on a feature until a stopping criterion is reached and hence, that particular node is not split further. Such a leaf node is assigned a label computed from the associated dataset. In our case, we associate two labels with each node: First, a regression label $\widehat{\vec{y}}^{\mathit{regression}}_{\mathcal{D}} \in \mathbb{R}^k$ obtained by averaging the labels in the associated dataset $\mathcal{D}$ and second, a ranking label $\widehat{\vec{y}}^{\mathit{ranking}}_{\mathcal{D}} \in \mathbb{R}^k$ obtained by computing a consensus ranking through Borda's method \autocite{lin2010rank}. 

We choose splits, consisting of a feature $f^* \in \mathbb{F}$, where $\mathbb{F}$ is the set of features, and a split point $p^*$, to minimize the weighted sum of the resulting dataset's losses wrt. the corresponding node labels, i.e.
\begin{equation}\label{eq:split_optimization_problem}
    (f^*,p^*) \in \argmin\limits_{(f,p) \in \mathbb{F}\times\mathbb{R}} \frac{|\dataset_{train}^+|}{|\dataset_{train}|} \cdot \nodewiseLoss(\mathcal{D^+}) + \frac{|\dataset_{train}^-|}{|\dataset_{train}|} \cdot \nodewiseLoss(\mathcal{D^-}) \,\,\, .
\end{equation}

These losses quantify the homogeneity of labels in the dataset and are calculated as a convex combination of ranking and regression losses $\nodewiseLoss(\dataset) = \lambda \nodewiseLoss_{ranking}(\dataset) + (1 - \lambda) \nodewiseLoss_{regression}(\dataset)$
where $\nodewiseLoss(\mathcal{D}) = \frac{1}{\vert \mathcal{D} \vert} \sum\nolimits_{(\vec{x}_I, \vec{y}_I) \in \mathcal{D}} \ell(\vec{y}_I, \widehat{\vec{y}}_{\mathcal{D}})$ and $\widehat{\vec{y}}_{\mathcal{D}}$ either corresponds to the ranking or regression label depending on whether $\ell$ is a ranking or regression loss function. We solve the optimization problem in \autoref{eq:split_optimization_problem} by a simple enumeration of all possible features and splitting points imposed by the training data and choosing the best one. We utilize the mean squared error over all algorithms and instances as a regression loss $\nodewiseLoss_{regression}$ as in \cite{jonas_combined_ranking_and_regerssion}. As a ranking loss, we leverage the Spearman correlation turned into a loss function by subtracting it from $1$, as we found this to work best in preliminary experiments. For the same reason we leverage the depth of a tree as a stopping criterion.

At prediction time, we propagate the instance down the tree until a leaf node $l$ with $\dataset_l$ is reached. Based on label $\widehat{\vec{y}}^{\mathit{regression}}_{\mathcal{D}_l}$ we finally return the algorithm performing best according to this label.

Since the choice of split is dependent on the utilized loss functions, their behavior is the dominant factor in the model's quality. However, we found that not all ranking loss functions are well suited for Hybrid Forests and a mismatch in the scale of ranking and regression losses can result in one loss dominating the other thereby mitigating the impact of $\lambda$. To solve this we scaled the losses to the unit interval by scaling the performance data and dividing the ranking loss by the maximum possible loss. Moreover,  the performance of \name heavily depends on the right choice of $\lambda$.

\section{Evaluation}
We assess the quality of HARRIS with an experimental evaluation on a small subset of the ASLib benchmark \autocite{AsLib}. All experiments were run on Intel Xeon E5-2695 v3 @ 2.30GHz CPU and 64 GB RAM. To set our results into context, we evaluate against ISAC \autocite{isac}, random forest regressor (RFR) that predicts each algorithms performance with a random forests, and SATzilla'11 \autocite{satzilla} as done in several recent works \cite{RunToSurvive,metaas,tornede2022algorithm}. In the interest of reproducibility, all code is available at \footnote{Github link: https://github.com/LukasFehring/HARRIS-Hybrid\_rAnking\_and\_RegRessIon\_foreSts}.

The quality of each approach is evaluated using 10-fold cross validation with Kendall's Tau-b \autocite{kendall1945treatment} and PAR10 \autocite{AsLib}. Kendall's Tau quantifies the correlation between two rankings, where $1$ indicates a perfect and $-1$ an inverse correlation. The PAR10 score corresponds to the runtime of the selected algorithm, if it is below a threshold $C$ and $10 \cdot C$ otherwise. This threshold $C$ is provided by the benchmark and corresponds to an upper bound on the runtime. 

\begin{table}[h]
    \centering
    \caption{Quality of the best known \name configuration and competitors quantified with PAR10.}
    \resizebox{0.65\textwidth}{!}{%
    \begin{tabular}{l!{\vrule width 1pt}rl|rl|rl|rl}
 &  \multicolumn{2}{c|}{\multirow{2}{*}{HARRIS}} & \multicolumn{2}{c|}{\multirow{2}{*}{ISAC}}&  \multicolumn{2}{c|}{\multirow{2}{*}{RFR}}  & \multicolumn{2}{c}{\multirow{2}{*}{SAT}} \\
 Scenario Name & & & & & & & & \\
\Xhline{1pt}
CSP-Minizinc-Time-2016 &    \textbf{476.97}&$\pm$661.60 &   1194.64&$\pm$592.74 &   1044.55&$\pm$886.96 &  1058.08&$\pm$1184.75 \\
MIP-2016               &  \textbf{1728.82}&$\pm$1649.62 &  2975.35&$\pm$3205.29 &  4332.53&$\pm$3320.56 &  2989.38&$\pm$2836.52 \\
QBF-2016               &   \textbf{1382.08}&$\pm$328.42 &   1704.74&$\pm$757.74 &   1722.20&$\pm$836.78 &   1607.81&$\pm$627.32 \\
CPMP-2015              &  \textbf{4891.47}&$\pm$1205.64 &  6094.06&$\pm$1972.29 &  5634.73&$\pm$2181.76 &  5152.87&$\pm$1521.40 \\
ASP-POTASSCO           &     209.47&$\pm$59.07 &    348.57&$\pm$133.53 &     \textbf{178.81}&$\pm$52.20 &     236.48&$\pm$74.78 \\
MAXSAT12-PMS           &    795.44&$\pm$399.61 &   1067.84&$\pm$700.12 &    631.14&$\pm$425.60 &    \textbf{553.61}&$\pm$371.80 \\
QBF-2011               &   2464.69&$\pm$721.31 &  3271.56&$\pm$1270.76 &   1865.75&$\pm$804.27 &   \textbf{1520.36}&$\pm$630.32 \\
SAT12-HAND             &   2150.58&$\pm$497.06 &   2587.54&$\pm$484.89 &   1552.95&$\pm$264.20 &   \textbf{1135.70}&$\pm$204.81 \\
SAT12-ALL              &   2476.95&$\pm$202.07 &   1999.36&$\pm$321.40 &   \textbf{1144.46}&$\pm$280.86 &   1349.94&$\pm$173.25 \\
\hline
Average Rank           & \multicolumn{2}{c|}{2.11} & \multicolumn{2}{c|}{3.56} & \multicolumn{2}{c|}{2.33} & \multicolumn{2}{c}{\textbf{2.00}} \\
\end{tabular}%
}
\label{evaluation_table}
\end{table}

\autoref{evaluation_table} displays the PAR10 scores averaged across all folds of each approach on the corresponding scenario including the standard deviation. Bold letters indicate the best performance. Note that the performances shown for \name are optimistic as they correspond to the best performance achieved by varying $\lambda$ in steps of $0.1$ and the tree depth in $\{2,4,6,8.10\}$. Thus, they can only serve to get an idea of what \name is capable of, if $\lambda$ can be tuned correctly. According to the average rank, \name is the second best approach.

\begin{figure}[h] 
    \centering
    \includegraphics[width=.9\linewidth]{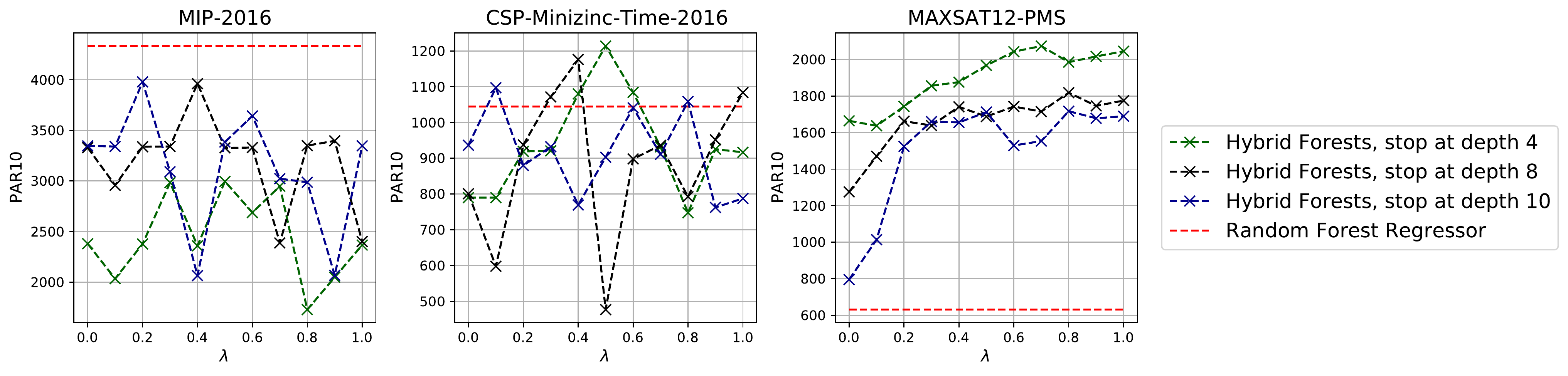}
    \caption{Visulisation of $\lambda$'s impact on the quality of HARRIS}
    \label{evaluation_figure}
\end{figure}

\autoref{evaluation_figure} visualizes the change in quality of \name with fixed depth for varying $\lambda$ in the PAR10 metric. The results indicate that while $\lambda$ strongly impacts the overall model quality, there are scenarios for which HARRIS is the superior/inferior model. More figures can be found in the appendix (\autoref{appendix}).

\section{Conclusion}
In this work, we proposed a hybrid ranking and regression tree-based approach to AS called, \name. Conceptually, \name alleviates the weaknesses of pure ranking and regression AS solutions. In a prototypical experimental study, we showed that with appropriately set hyperparameters, \name can outperform existing algorithm selectors on some scenarios. In future work, we plan to investigate whether tuning these hyperparameters automatically via means of hyperparameter optimization \cite{bischl2021hyperparameter} yields good values on a scenario as suggested in \cite{lindauerHHS15}. Moreover, we plan to investigate other options for combining regression and ranking loss functions, for example, by working with probabilistic loss functions as this alleviates possible problems related to different scales. 

\newpage
\ifthenelse{\boolean{blind}}
{
}
{
\subsection*{Acknowledgements}
This work was supported by the German Research Foundation (DFG) within the Collaborative Research Center ``On-The-Fly Computing'' (SFB 901/3 project no.\ 160364472).
}

\printbibliography

@article{satzilla,
  title={Hydra-MIP: Automated algorithm configuration and selection for mixed integer programming},
  author={Xu, Lin and Hutter, Frank and Hoos, Holger H and Leyton-Brown, Kevin},
  journal={RCRA workshop on experimental evaluation of algorithms for solving problems with combinatorial explosion@IJCAI 2011},
  pages={16--30},
  year={2011}
}

@article{rice_algorithm_selection,
  author    = {John R. Rice},
  title     = {The Algorithm Selection Problem},
  journal   = {Adv. Comput.},
  volume    = {15},
  pages     = {65--118},
  year      = {1976},
  url       = {https://doi.org/10.1016/S0065-2458(08)60520-3},
  doi       = {10.1016/S0065-2458(08)60520-3},
  bibsource = {dblp computer science bibliography, https://dblp.org}
}

@book{breimann_cart,
  author    = {Leo Breiman and
               J. H. Friedman and
               R. A. Olshen and
               C. J. Stone},
  title     = {Classification and Regression Trees},
  publisher = {Wadsworth},
  year      = {1984},
  isbn      = {0-534-98053-8},
  bibsource = {dblp computer science bibliography, https://dblp.org}
}

@inproceedings{jonas_combined_ranking_and_regerssion,
  author    = {Jonas Hanselle and
               Alexander Tornede and
               Marcel Wever and
               Eyke H{\"{u}}llermeier},
  title     = {Hybrid Ranking and Regression for Algorithm Selection},
  booktitle = {{KI} 2020: Proceedings of the 43rd German Conference
               on AI},
  series    = {Lecture Notes in Computer Science},
  volume    = {12325},
  pages     = {59--72},
  publisher = {Springer},
  year      = {2020},
  url       = {https://doi.org/10.1007/978-3-030-58285-2\_5},
  doi       = {10.1007/978-3-030-58285-2\_5},
  bibsource = {dblp computer science bibliography, https://dblp.org}
}

@article{SAT2020,
  author    = {Nils Froleyks and
               Marijn Heule and
               Markus Iser and
               Matti J{\"{a}}rvisalo and
               Martin Suda},
  title     = {{SAT} Competition 2020},
  journal   = {Artificial Intelligence},
  volume    = {301},
  pages     = {103572},
  year      = {2021},
  url       = {https://doi.org/10.1016/j.artint.2021.103572},
  doi       = {10.1016/j.artint.2021.103572},
  bibsource = {dblp computer science bibliography, https://dblp.org}
}

@article{SAT2018,
    author    = {Marijn J. H. Heule and
               Matti J{\"{a}}rvisalo and
               Martin Suda},
  title     = {{SAT} Competition 2018},
  journal   = {J. Satisf. Boolean Model. Comput.},
  volume    = {11},
  number    = {1},
  pages     = {133--154},
  year      = {2019},
  url       = {https://doi.org/10.3233/SAT190120},
  doi       = {10.3233/SAT190120},
  bibsource = {dblp computer science bibliography, https://dblp.org}
}

@inproceedings{CombinedRankingAndRegression,
  author    = {D. Sculley},
  title     = {Combined regression and ranking},
  booktitle = {{SIGKDD} 2010: Proceedings of the 16th {ACM} International Conference on
               Knowledge Discovery and Data Mining},
  pages     = {979--988},
  publisher = {{ACM}},
  year      = {2010},
  url       = {https://doi.org/10.1145/1835804.1835928},
  doi       = {10.1145/1835804.1835928},
  bibsource = {dblp computer science bibliography, https://dblp.org}
}

@article{AsLib,
  author    = {Bernd Bischl and
               Pascal Kerschke and
               Lars Kotthoff and
               Marius Lindauer and
               Yuri Malitsky and
               Alexandre Fr{\'{e}}chette and
               Holger H. Hoos and
               Frank Hutter and
               Kevin Leyton{-}Brown and
               Kevin Tierney and
               Joaquin Vanschoren},
  title     = {ASlib: {A} benchmark library for algorithm selection},
  journal   = {Artificial Intelligence},
  volume    = {237},
  pages     = {41--58},
  year      = {2016},
  url       = {https://doi.org/10.1016/j.artint.2016.04.003},
  doi       = {10.1016/j.artint.2016.04.003},
  bibsource = {dblp computer science bibliography, https://dblp.org}
}

@inproceedings{isac,
  author    = {Serdar Kadioglu and
               Yuri Malitsky and
               Meinolf Sellmann and
               Kevin Tierney},
  title     = {{ISAC} - Instance-Specific Algorithm Configuration},
  booktitle = {{ECAI} 2010: Proceedings of the 19th European Conference on Artificial Intelligence},
  series    = {Frontiers in Artificial Intelligence and Applications},
  volume    = {215},
  pages     = {751--756},
  publisher = {{IOS} Press},
  year      = {2010},
  url       = {https://doi.org/10.3233/978-1-60750-606-5-751},
  doi       = {10.3233/978-1-60750-606-5-751},
  bibsource = {dblp computer science bibliography, https://dblp.org}
}

@inproceedings{RunToSurvive,
  author    = {Alexander Tornede and
               Marcel Wever and
               Stefan Werner and
               Felix Mohr and
               Eyke H{\"{u}}llermeier},
  title     = {Run2Survive: {A} Decision-theoretic Approach to Algorithm Selection
               based on Survival Analysis},
  booktitle = {{ACML} 2020: Proceedings of The 12th Asian Conference on Machine Learning},
  series    = {Proceedings of Machine Learning Research},
  volume    = {129},
  pages     = {737--752},
  publisher = {{PMLR}},
  year      = {2020},
  url       = {http://proceedings.mlr.press/v129/tornede20a.html},
  bibsource = {dblp computer science bibliography, https://dblp.org}
}

@inproceedings{xu2007satzilla,
  author    = {Lin Xu and
               Frank Hutter and
               Holger H. Hoos and
               Kevin Leyton{-}Brown},
  title     = {The Design and Analysis of an Algorithm Portfolio for {SAT}},
  booktitle = {{CP} 2007: Proceedings of the 13th International Conference on Constraint Programming},
  series    = {Lecture Notes in Computer Science},
  volume    = {4741},
  pages     = {712--727},
  publisher = {Springer},
  year      = {2007},
  url       = {https://doi.org/10.1007/978-3-540-74970-7\_50},
  doi       = {10.1007/978-3-540-74970-7\_50},
  bibsource = {dblp computer science bibliography, https://dblp.org}
}

@inproceedings{tornedeWH20,
  author    = {Alexander Tornede and
               Marcel Wever and
               Eyke H{\"{u}}llermeier},
  title     = {Extreme Algorithm Selection with Dyadic Feature Representation},
  booktitle = {{DS} 2020: Proceedings of the 23rd International Conference on Discovery Science},
  series    = {Lecture Notes in Computer Science},
  volume    = {12323},
  pages     = {309--324},
  publisher = {Springer},
  year      = {2020},
  url       = {https://doi.org/10.1007/978-3-030-61527-7\_21},
  doi       = {10.1007/978-3-030-61527-7\_21},
  bibsource = {dblp computer science bibliography, https://dblp.org}
}

@inproceedings{brazdil2000comparison,
  author    = {Pavel Brazdil and
               Carlos Soares},
  title     = {A Comparison of Ranking Methods for Classification Algorithm Selection},
  booktitle = {{ECML} 2000: Proceedings of the 11th European Conference on Machine
               Learning},
  series    = {Lecture Notes in Computer Science},
  volume    = {1810},
  pages     = {63--74},
  publisher = {Springer},
  year      = {2000},
  url       = {https://doi.org/10.1007/3-540-45164-1\_8},
  doi       = {10.1007/3-540-45164-1\_8},
  bibsource = {dblp computer science bibliography, https://dblp.org}
}

@inproceedings{cunha2018label,
  author    = {Tiago Cunha and
               Carlos Soares and
               Andr{\'{e}} C. P. L. F. de Carvalho},
  title     = {A label ranking approach for selecting rankings of collaborative filtering
               algorithms},
  booktitle = {{SAC} 2018: Proceedings of the 33rd Annual {ACM} Symposium on Applied Computing},
  pages     = {1393--1395},
  publisher = {{ACM}},
  year      = {2018},
  url       = {https://doi.org/10.1145/3167132.3167418},
  doi       = {10.1145/3167132.3167418},
  bibsource = {dblp computer science bibliography, https://dblp.org}
}

@article{lin2010rank,
  title={Rank aggregation methods},
  author={Lin, Shili},
  journal={Wiley Interdisciplinary Reviews: Computational Statistics},
  volume={2},
  number={5},
  pages={555--570},
  year={2010},
  publisher={Wiley Online Library}
}

@book{spearman_general_1961,
	address = {East Norwalk, CT, US},
	series = {Studies in individual differences: {The} search for intelligence},
	title = {"{General} {Intelligence}" {Objectively} {Determined} and {Measured}},
	publisher = {Appleton-Century-Crofts},
	author = {Spearman, C.},
	year = {1961},
	doi = {10.1037/11491-006},
}

@article{bischl2021hyperparameter,
  author    = {Bernd Bischl and
               Martin Binder and
               Michel Lang and
               Tobias Pielok and
               Jakob Richter and
               Stefan Coors and
               Janek Thomas and
               Theresa Ullmann and
               Marc Becker and
               Anne{-}Laure Boulesteix and
               Difan Deng and
               Marius Lindauer},
  title     = {Hyperparameter Optimization: Foundations, Algorithms, Best Practices
               and Open Challenges},
  journal   = {CoRR},
  volume    = {abs/2107.05847},
  year      = {2021},
  url       = {https://arxiv.org/abs/2107.05847},
  eprinttype = {arXiv},
  eprint    = {2107.05847},
  bibsource = {dblp computer science bibliography, https://dblp.org}
}

@inproceedings{hanselleTWH21,
  author    = {Jonas Hanselle and
               Alexander Tornede and
               Marcel Wever and
               Eyke H{\"{u}}llermeier},
  title     = {Algorithm Selection as Superset Learning: Constructing Algorithm Selectors
               from Imprecise Performance Data},
  booktitle = {{PAKDD} 2021: Proceedings of the 25th Pacific-Asia
               Conference},
  series    = {Lecture Notes in Computer Science},
  volume    = {12712},
  pages     = {152--163},
  publisher = {Springer},
  year      = {2021},
  url       = {https://doi.org/10.1007/978-3-030-75762-5\_13},
  doi       = {10.1007/978-3-030-75762-5\_13},
  bibsource = {dblp computer science bibliography, https://dblp.org}
}

@article{lindauerHHS15,
  author    = {Marius Lindauer and
               Holger H. Hoos and
               Frank Hutter and
               Torsten Schaub},
  title     = {AutoFolio: An Automatically Configured Algorithm Selector},
  journal   = {Journal of Artificial Intelligence Research},
  volume    = {53},
  pages     = {745--778},
  year      = {2015},
  url       = {https://doi.org/10.1613/jair.4726},
  doi       = {10.1613/jair.4726},
  bibsource = {dblp computer science bibliography, https://dblp.org}
}

@article{kendall1945treatment,
  title={The treatment of ties in ranking problems},
  author={Kendall, Maurice G},
  journal={Biometrika},
  volume={33},
  number={3},
  pages={239--251},
  year={1945},
  publisher={JSTOR}
}

@article{tornede2022algorithm,
  title={Algorithm selection on a meta level},
  author={Tornede, Alexander and Gehring, Lukas and Tornede, Tanja and Wever, Marcel and H{\"u}llermeier, Eyke},
  journal={Machine Learning},
  pages={1--34},
  year={2022},
  publisher={Springer}
}

@inproceedings{metaas,
  title={Towards Meta-Algorithm Selection},
  author={Tornede, A. and Wever, M. and H{\"u}llermeier, E.},
  booktitle={{Workshop on Meta-Learning (MetaLearn 2020) @ NeurIPS 2020}},
  year={2020}
}

@article{breiman01,
  author    = {Leo Breiman},
  title     = {Random Forests},
  journal   = {Mach. Learn.},
  volume    = {45},
  number    = {1},
  pages     = {5--32},
  year      = {2001},
  url       = {https://doi.org/10.1023/A:1010933404324},
  doi       = {10.1023/A:1010933404324},
  bibsource = {dblp computer science bibliography, https://dblp.org}
}

\newpage
\section{Appendix}\label{appendix}
\subsection*{Benchmark Scenarios}
As mentioned in the paper, we evaluated the competitors performances with the ASlib \autocite{AsLib} benchmark. However, we were not able to evaluate on all scenarios but just a subset of them. Key properties of them are shown in \autoref{scenario_properties}.

\begin{table}[h]
    \centering
    \caption{Properties of the benchmark scenarios used for model evaluation.}
    \resizebox{\textwidth}{!}{
\begin{tabular}{l|cccccccccc}
               Scenario &   Problem &  Instances &  Algorithms &  Features &  Unsolved Instances &  Proportion Unsolved Instances &  Proportion Missing Evaluation &       Cutoff\\
\hline
           ASP-POTASSCO &       ASP &       1294 &          11 &       138 &                  82 &                         0.06 &                 0.20 &        600.0\\
              CPMP-2015 &      CPMP &        527 &           4 &        22 &                   0 &                         0.00 &                 0.28 &       3600.0\\
 CSP-Minizinc-Time-2016 &       CSP &        100 &          20 &        95 &                  17 &                         0.17 &                 0.50 &       1200.0\\
           MAXSAT12-PMS &  MAXSAT12 &        876 &           6 &        37 &                 129 &                         0.15 &                 0.41 &       2100.0\\
               MIP-2016 &       MIP &        218 &           5 &       143 &                   0 &                         0.00 &                 0.20 &       7200.0\\
               QBF-2011 &       QBF &       1368 &           5 &        46 &                 314 &                         0.23 &                 0.55 &       3600.0\\
               QBF-2016 &       QBF &        825 &          24 &        46 &                  55 &                         0.07 &                 0.36 &       1800.0\\
             SAT12-HAND &     SAT12 &        767 &          31 &       115 &                 229 &                         0.30 &                 0.67 &       1200.0\\
             SAT12-INDU &     SAT12 &       1167 &          31 &       115 &                 209 &                         0.18 &                 0.50 &       1200.0\\
\end{tabular}}
\label{scenario_properties}
\end{table}

An instance is unsolved if no candidate algorithm solves the instance before the cutoff is reached. An evaluation of some algorithm on an instance is missing if the algorithm does not finish it's calculation before the cutoff is reached.

\subsection*{Further Evaluation Results}
In the paper we were only able to give a brief overview over the results of our evalaution. Further resutls are shown in the following figures.

\autoref{all_scenarios_par10} shows the results of our PAR10 evaluation for all considered scenarios.

\begin{figure}[h]
    \centering
    \includegraphics[height=.47\textheight]{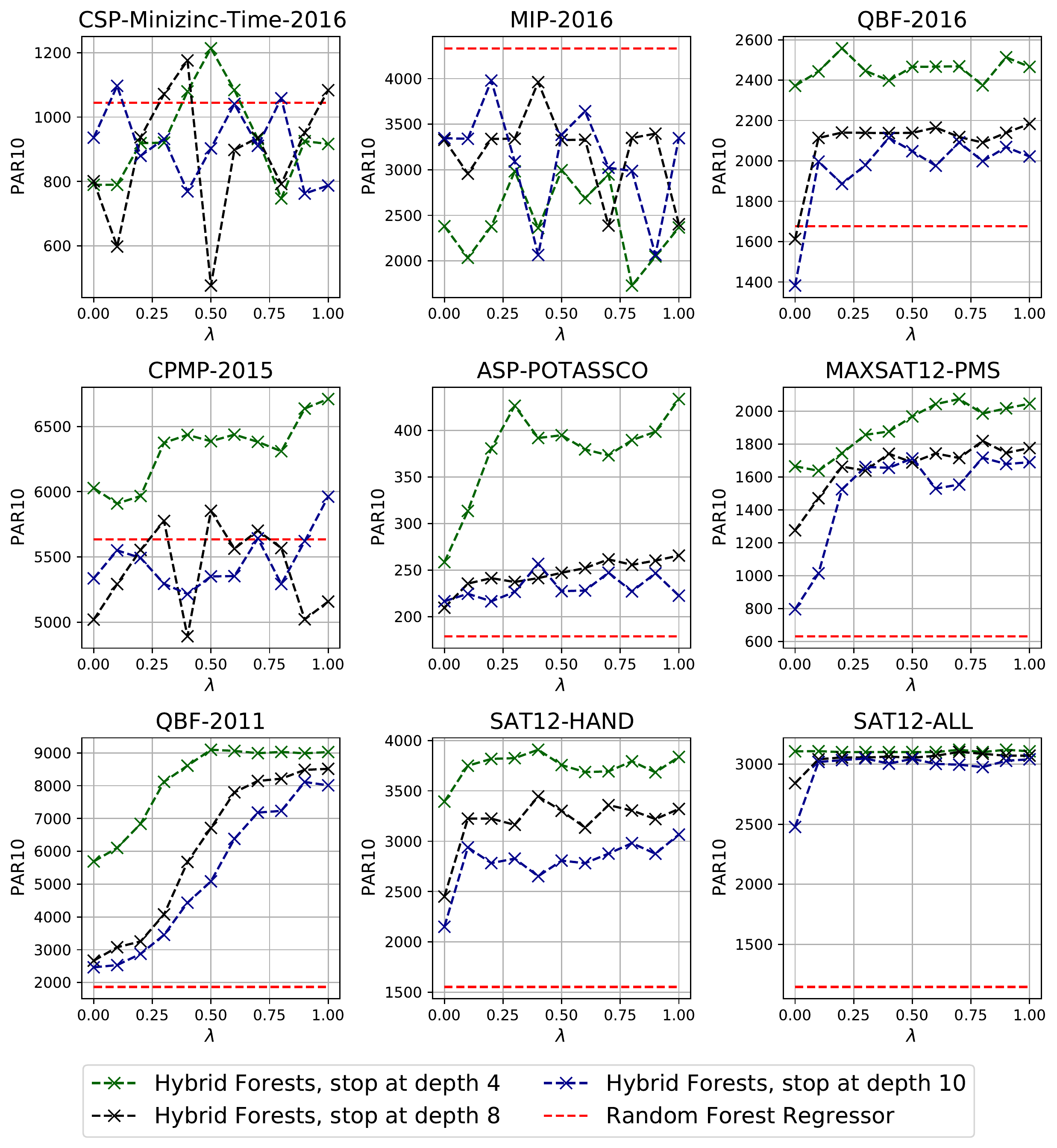}
    \caption{Quality Comparison of different HARRIS configurations with the Random Forest Regressor. The model quality is quantified with PAR10.}
    \label{all_scenarios_par10}
\end{figure}

\begin{figure}[h]
    \centering
    \includegraphics[height=.37\textheight]{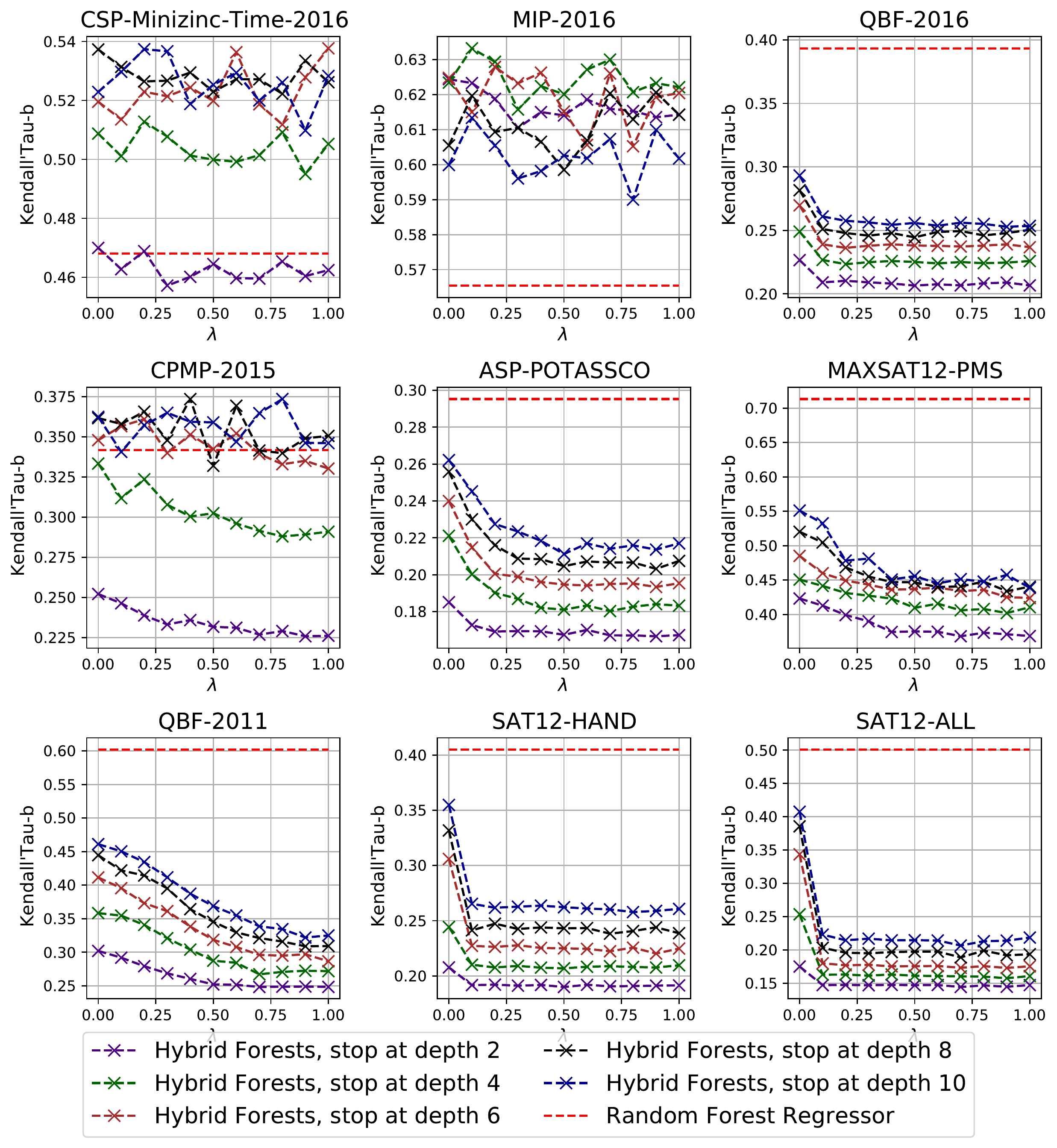}
    \caption{Quality Comparison of different HARRIS combinations with the Random Forest Regressor. The model quality is quantified with the Kendall's Tau metric.}
    \label{all_scenarios_kendalls}
\end{figure}

\autoref{all_scenarios_kendalls} shows the results of our Kendall's Tau-b evaluation for all considered scenarios.

\begin{figure}[h]
    \centering
    \includegraphics[height = .37\textheight]{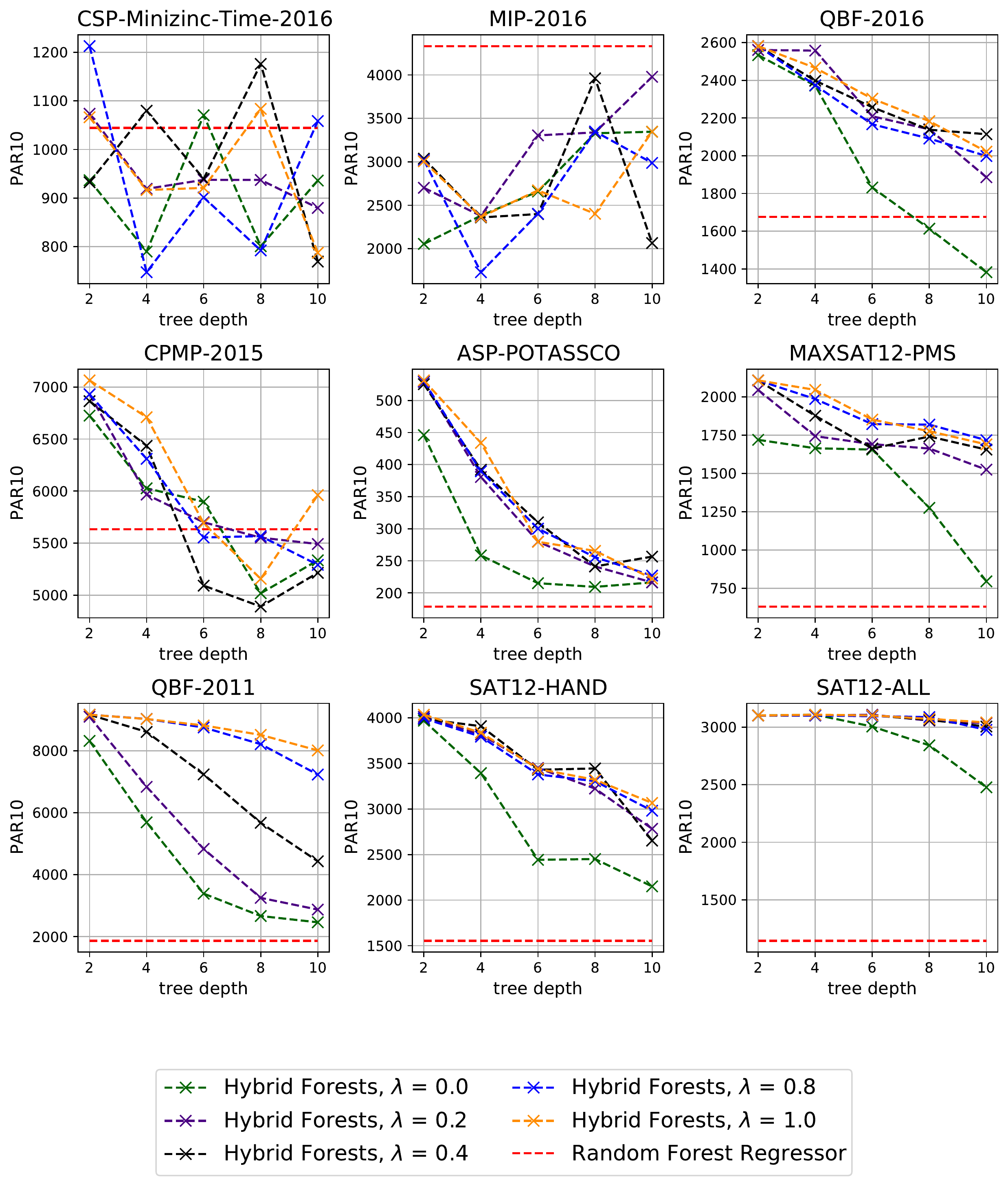}
    \caption{Evaluation of the stopping criterion's impact on the overall model quality. Note that the results indicate that \name might improve for increasing depth on some scenarios.}
    \label{depth_evaluation}
\end{figure}

\autoref{depth_evaluation} shows the results of our evaluation of different tree depths in terms of the PAR10 number of the resulting algorithm selector.
\end{document}